\documentclass[letterpaper]{article} 
\usepackage{aaai2026}  
\usepackage{times}  
\usepackage{helvet}  
\usepackage{courier}  
\usepackage[hyphens]{url}  
\usepackage{graphicx} 
\usepackage{natbib}  
\usepackage{caption} 
\usepackage{algorithm}
\usepackage{algorithmic}

\usepackage{amsmath}
\usepackage{amssymb}
\usepackage{amsthm}

\newtheorem{theorem}{Theorem}
\newtheorem{lemma}{Lemma}
\newtheorem{corollary}{Corollary}
\newtheorem{assumption}{Assumption}
\newtheorem{definition}{Definition}
\newtheorem{remark}{Remark}

\usepackage{newfloat}
\usepackage{listings}
\DeclareCaptionStyle{ruled}{labelfont=normalfont,labelsep=colon,strut=off} 
\floatstyle{ruled}
\newfloat{listing}{tb}{lst}{}
\floatname{listing}{Listing}
\title{Optimal Look-back Horizon for Time Series Forecasting in Federated Learning}
\author{
    Dahao Tang\textsuperscript{\rm 1},
    Nan Yang\textsuperscript{\rm 1}
    \thanks{Corresponding authors.},
    Yanli Li\textsuperscript{\rm 1},
    Zhiyu Zhu\textsuperscript{\rm 2},
    Zhibo Jin\textsuperscript{\rm 2},
    Dong Yuan\textsuperscript{\rm 1}
    \footnotemark[1]
}
\affiliations{
    \textsuperscript{\rm 1}University of Sydney \\
    \textsuperscript{\rm 2}University of Technology Sydney

    dahao.tang@sydney.edu.au,
    n.yang@sydney.edu.au,
    yanli.li@sydney.edu.au,
    zhiyu.zhu@student.uts.edu.au,
    zhibo.jin@student.uts.edu.au,
    dong.yang@sydney.edu.au
}

\begin{document}

\maketitle

\begin{abstract}

Selecting an appropriate look-back horizon remains a fundamental challenge in time series forecasting (TSF), particularly in the federated learning scenarios where data is decentralized, heterogeneous, and often non-independent. While recent work has explored horizon selection by preserving forecasting-relevant information in an intrinsic space, these approaches are primarily restricted to centralized and independently distributed settings. This paper presents a principled framework for adaptive horizon selection in federated time series forecasting through an intrinsic space formulation. We introduce a synthetic data generator (SDG) that captures essential temporal structures in client data, including autoregressive dependencies, seasonality, and trend, while incorporating client-specific heterogeneity. Building on this model, we define a transformation that maps time series windows into an intrinsic representation space with well-defined geometric and statistical properties. We then derive a decomposition of the forecasting loss into a Bayesian term, which reflects irreducible uncertainty, and an approximation term, which accounts for finite-sample effects and limited model capacity. Our analysis shows that while increasing the look-back horizon improves the identifiability of deterministic patterns, it also increases approximation error due to higher model complexity and reduced sample efficiency. We prove that the total forecasting loss is minimized at the smallest horizon where the irreducible loss starts to saturate, while the approximation loss continues to rise. This work provides a rigorous theoretical foundation for adaptive horizon selection for time series forecasting in federated learning.

\end{abstract}

\section{Introduction}

Time series forecasting (TSF) underpins numerous high-impact domains, including finance \cite{zivot2006modeling}, healthcare \cite{futoma2017learning}, and energy systems \cite{kong2019short}, where accurate prediction of future values from historical trends is crucial for informed decision-making and operational efficiency. A central modeling choice in TSF is the selection of the look-back horizon, defined as the number of past time steps used as input. This choice significantly influences model complexity, predictive accuracy, and generalization performance \cite{lim2020temporal}.

Traditionally, the look-back horizon is treated as a tunable hyperparameter, often selected via cross-validation or heuristic search. Recent theoretical advances offer a more principled perspective. Shi et al. \cite{shi2024scaling} propose a scaling law theory based on a theoretical framework that embeds time series into an intrinsic representation space, allowing the forecasting loss to be decomposed into two components: Bayesian error, capturing irreducible uncertainty from noise and limited information, and approximation error, reflecting the model’s capacity to learn the true mapping. This decomposition enables analytical reasoning about the optimal look-back horizon as a function of dataset size, model complexity, and intrinsic dimensionality \cite{sharma2020scaling,bahri2024explaining}. Empirical results support these insights, showing that the optimal horizon grows with data availability and varies by model type. For example, channel-dependent models like iTransformer \cite{liu2024itransformer} benefit from shorter horizons under limited data, while linear models such as NLinear \cite{zeng2022transformers} maintain performance with longer horizons due to smoother feature decay and lower intrinsic complexity \cite{xu2024fits,toner2024analysis}.

However, this framework relies on strong assumptions, including centralized data, independent identically distribution (IID), and homogeneous model architectures, which are often violated in real-world federated learning scenarios. In such decentralized settings, data is distributed across clients with diverse distributions, sequence lengths, and domain characteristics \cite{kairouz2021advances}. Applying a globally fixed horizon in this context may lead to mismatches between local dynamics and model inputs, degrading forecasting performance \cite{edwards2024scaling}. Moreover, real-world data frequently exhibits feature sparsity, variable noise levels, and heterogeneous scaling behaviors, challenging the smooth manifold assumptions in intrinsic dimension theory \cite{levi2024scaling,zador1982quantization}. These limitations highlight the need for adaptive horizon strategies that account for both data heterogeneity and localized model constraints. Integrating hybrid architectures or meta-learning mechanisms with principled theoretical foundations, such as those introduced by Shi et al., presents a promising direction for addressing these challenges in federated time series forecasting.

This paper addresses the challenge of selecting the optimal look-back horizon for time series forecasting in federated learning environments characterized by non-IID client data. We develop a principled framework that leverages a structured Synthetic Data Generator (SDG) to model core temporal patterns (e.g., autoregressive dynamics, seasonality, and trends), while capturing client-specific heterogeneity. Using the SDG as a foundation, we construct a data-aware transformation that maps time series windows into an intrinsic representation space with well-defined geometric and statistical properties. This enables a rigorous loss decomposition into irreducible (Bayesian) and approximation components, each tied to the underlying generative structure. Crucially, the formulation reveals how the informativeness of historical context and thus the optimal look-back horizon varies across clients depending on their local dynamics and data regimes. Our analysis shows that the total forecasting loss is minimized at the smallest horizon where the Bayesian error saturates and the approximation error begins to dominate, yielding a theoretically grounded, client-adaptive criterion for horizon selection in federated forecasting settings. Our contributions include the following:

\begin{itemize}
    \item We propose a novel intrinsic space formulation that transforms heterogeneous, non-IID multivariate time series into a compact and geometry-preserving representation. This space is rigorously characterized by bi-Lipschitz continuity, intrinsic dimensionality saturation, and inter-horizon compatibility, enabling consistent comparison and reasoning across clients and temporal contexts.
    \item We establish a tight decomposition of predictive loss into irreducible (Bayesian) and approximation components, each analytically tied to the structural elements of time series data (e.g., AR memory, seasonality, trend) and the look-back horizon. Our analysis uncovers the fundamental bias–variance trade-off that governs forecasting performance in federated settings.
    \item We prove that the total loss is unimodal with respect to the horizon length and identify the smallest sufficient horizon as its global minimizer. This result provides the first rigorous criterion for horizon selection in time series forecasting and introduces a new design principle for model construction under sample-limited, heterogeneous environments.
\end{itemize}

\section{Related Work} 

\subsection{Horizon Selection and Intrinsic Representation}

A central yet understudied question in time series forecasting (TSF) is how much historical context, i.e., look-back horizon, is truly needed for accurate prediction \cite{kim2025comprehensive}. Traditional statistical models such as ARIMA select lag length using information criteria like AIC \cite{akaike1974new,box2015time}, which implicitly perform horizon selection under strong linearity assumptions. While interpretable, these methods struggle to capture nonlinear or long-range dependencies.
Modern deep learning approaches, including LSTMs \cite{hochreiter1997long}, hybrid models like LSTNet \cite{lai2018modeling}, and attention-based architectures such as the Temporal Fusion Transformer \cite{lim2020temporal} or Informer \cite{zhou2021informer}, have greatly improved modeling capacity. However, they still treat the input horizon as a tunable hyperparameter, typically set through validation or heuristics, without theoretical grounding. This empirical approach can lead to overfitting, underfitting, or inefficient use of data, particularly in settings with limited or distributed samples \cite{woo2023deeptime,koparanov2020lookback}.

Recent theoretical work has begun to formalize the horizon selection problem. Notably, Shi et al. (2024) analyze how forecasting error scales with input length, dataset size, and model complexity, revealing a trade-off: longer horizons can improve identifiability of temporal structure but also increase approximation error due to model limitations and finite data \cite{shi2024scaling}. Their framework introduces the notion of an intrinsic representation space, where the forecasting loss decomposes into two parts: a Bayesian (irreducible) error reflecting inherent unpredictability, and an approximation error arising from statistical and model constraints. This idea builds on Takens' embedding theorem \cite{takens2006detecting}, which implies that a system’s future behavior can be reconstructed from a finite number of past observations, defining an intrinsic dimension sufficient for prediction. However, Shi's work assumes centralized and IID data, limiting its relevance to modern federated learning scenarios where data is distributed, non-IID, and client-specific.
We extend this theory to federated, non-IID settings by introducing an intrinsic representation that captures essential temporal structure across clients. This enables a principled approach to selecting the optimal look-back horizon in decentralized forecasting.

\subsection{Time Series Forecasting in Federated Learning}

Federated learning (FL) enables decentralized training across clients without sharing raw data. The foundational FedAvg algorithm introduced by McMahan et al. (2017) laid the groundwork for collaborative model training in privacy-sensitive environments \cite{mcmahan2017communication}. However, FL under non-IID data poses major challenges, including model divergence, degraded generalization, and client imbalance.
To address data heterogeneity, methods like FedProx \cite{li2020federated} introduce regularization terms that stabilize optimization across diverse client distributions. In time series forecasting specifically, recent works apply FL to real-world sequential tasks, such as traffic and energy demand prediction, but focus primarily on model architecture and aggregation \cite{perifanis2023federated}.
These systems rarely examine how temporal structure varies across clients or how such variation affects forecasting horizons. While personalization and communication efficiency have been explored, no prior work provides a theoretical framework for look-back horizon selection in federated TSF. Our paper addresses this gap by analyzing horizon choice through the lens of synthetic modeling and intrinsic representation under client heterogeneity.

\section{Preliminary}
\label{sec:preliminary}

In this section, we defined the basic settings for time series forecasting in the federated learning scenario. More specifically, we propose a synthetic data generator (SDG) that well describes real-world non-IID data and implement a step-by-step transformation that converts the time series data described by the SDG into an intrinsic space that represents the information carried by a time series.

\subsection{Time Series Forecasting in Federated Learning}

We study $S$-step forecasting from a length-$H$ look-back window in a federated setting with $K$ clients. Client $k \in \{1, \dots, K\}$ holds a multivariate time series $\{x_{t}^{(k)}\}_{t=1}^{L_k}$ with $F$ features, $x_{t}^{(k)} \in \mathbb{R}^{F}$. For time index $t \in \{H, \dots, L_k - S\}$, define the input window and $S$-step target block as:
\begin{equation}
    X_{t,k}^{(H)} = \big[ x^{(k)}_{t-H+1}, \dots, x^{(k)}_{t} \big] \in \mathbb{R}^{F \times H},
\end{equation}
\begin{equation}
    Y_{t,k}^{(S)} = \big[ x^{(k)}_{t+1}, \dots, x^{(k)}_{t+S} \big] \in \mathbb{R}^{F \times S}.
\end{equation}

Training proceeds in rounds via standard FL aggregation (e.g., FedAvg). On client $k$, overlapping windows yield $D_k$ training samples; due to overlap, the number of effectively independent samples scales as $D_k / H$.

\subsection{Intrinsic Space Formulation}
\label{sec:intrinsic_space}

We adopt the concept of intrinsic space to represent the information carried by a time series. The intrinsic dimension $d_I$ of the intrinsic space is defined as the minimum number of dimensions required to represent the time series without losing significant information. To gain a deeper understanding of the intrinsic space, we investigate the typical structure of non-IID time series data in the federated learning scenario and propose a synthetic data generator that is both theoretically and empirically proven to be sound in describing the structure of the focused non-IID time series data.

\subsubsection{Synthetic Data Generator}
\label{sec:sdg}

A Synthetic Data Generator (SDG) is a parametric model designed to simulate univariate time series data, which often exhibits structural patterns characterized by seasonality, temporal dependence (AR memory), and trend \cite{kim2025comprehensive}.

For a given client $k$, feature $f$, and time step $t$, the synthetic observation $\hat{x}_{f,t,k}$ is defined as:
\begin{equation}
\label{eq:sdg}
    \begin{split}
        \hat{x}_{f,t,k} 
            &= \text{Seasonal}(A_{f,j,k}, T_{f,j,k}, \Theta_{f,j,k}) + \text{AR}_{p,k}(\phi_k) \\
            & \quad + \text{Trend}(\beta_{f,k}) + \epsilon_{f,t,k} \\
            &= \sum_{j=1}^{J} A_{f,j,k} \cdot \sin \left( \frac{2 \pi t}{T_{f,j,k}} + \theta_{f,j,k} \right) \\
            & \quad + \sum_{i=1}^{p} \phi_{k,i} \ x_{f,t-i,k} + \beta_{f,k} \ t + \epsilon_{f,t,k}.
    \end{split}
\end{equation}
Here, seasonality is represented by a sum of sinusoids, parameterized by amplitude $A_{f,j,k}$, period $T_{f,j,k}$, and phase shift $\theta_{f,j,k}$. Temporal dependence is modeled via an autoregressive process $\text{AR}_{p,k}(\phi_k) = \sum_{i=1}^{p} \phi_{k,i} \ x_{f,t-i,k}$, where $\phi_{k,i}$ are the lag coefficients specific to client $k$. The trend is captured by a linear component $\text{Trend}(\beta_{f,k}) = \beta_{f,k} \ t$. The additive noise term is drawn from a Gaussian distribution: $\epsilon_{f,t,k} \sim \mathcal{N}(\mu_{f,k}, \ \sigma_{f,k}^2)$.

\begin{figure}[t]
\centering
\includegraphics[width=0.99\columnwidth]{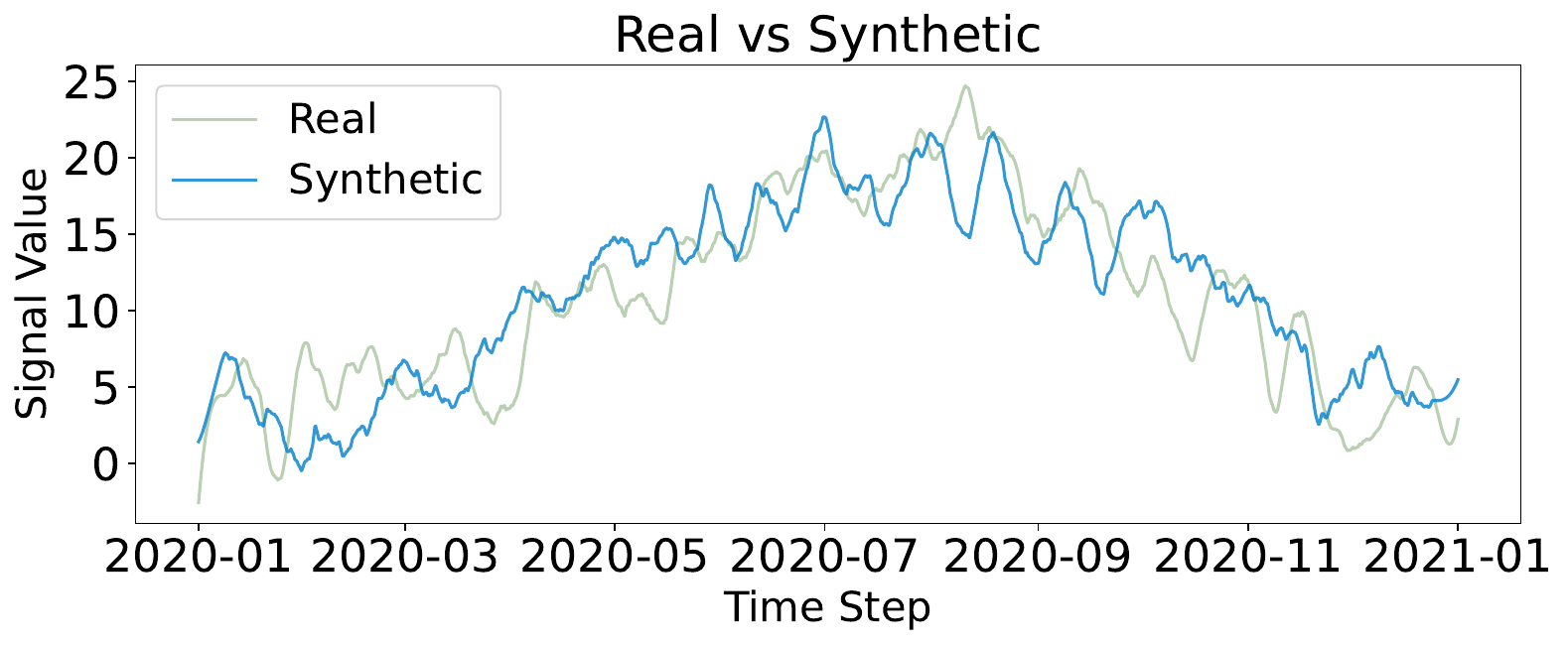} 
\caption{Comparison between real-world data and data generated by the SDG. The close alignment indicates that the SDG effectively captures the patterns present in real data.}
\label{fig1}
\end{figure}

We also provide empirical studies to demonstrate the validity of the SDG, as illustrated in Figure \ref{fig1}. Please refer to the Extended Version for more details.

\subsubsection{Feature Skewness Formulation}

In the federated learning scenario, each client tends to observe a different distribution of the same features in time series data, simulating feature skew \cite{wu2024spatio}. We apply a customized skewness partitioning method to create feature heterogeneity.

To be more precise, we construct an aﬀine transformation for each data point of the SDG: for client $k$, feature $k$:
\begin{equation}
    x_{f,t,k} = \Lambda_{f,k} \tilde{x}_{f,t,k} + \delta_{f,k}
\end{equation}
where $\Lambda_{fk}$ is the linear scale, which controls how the variance of the feature $f$, $\sigma_f^2$, changes for client $k$; $\delta_{fk}$ is the mean shift, which changes the mean of the feature $f$, $\mu_f$, for the client $k$. Note that, though the univariate SDG is able to describe each feature, each client is allowed to observe a subset of all the features.

\subsubsection{Intrinsic Space Construction}

At a high level, we construct a geometry-aware representation space that captures the essential temporal structure of non-IID time series through a transformation grounded in the SDG, which explicitly models autoregressive dependencies, seasonal cycles, and linear trends, and serves as a unifying scaffold for both analytical reasoning and empirical evaluation across heterogeneous clients.

Our construction is supported by a set of structural assumptions. These include: (i) compactness of the intrinsic image to ensure bounded representation norms; (ii) bi-Lipschitz continuity to preserve distances and guarantee stable inverses; (iii) a horizon-indexed intrinsic dimension that increases monotonically and saturates once all relevant temporal structure is captured; (iv) compatibility of representations across horizons via stable linear projections; (v) approximate commutativity between truncation and projection, ensuring robustness under input length variation; and (vi) a power-law spectrum of the intrinsic covariance, which enables efficient dimensionality reduction. These assumptions reflect statistical regularities commonly observed in time series data and enable a clean separation between modeling complexity and representational geometry.

The transformation pipeline proceeds in five steps: (1) \emph{Client-wise normalization} to remove affine feature skew and align marginal distributions; (2) \emph{Window flattening} to convert each normalized time-series segment into a fixed-length vector; (3) \emph{Global covariance estimation and eigendecomposition} to identify dominant axes of variation; (4) \emph{Intrinsic dimension estimation} based on the SDG and empirical spectrum; and (5) \emph{Projection into intrinsic space} via principal components. Specifically, the intrinsic dimension for client $k$ is approximated as:
\begin{equation}
   d_{I,k}(H) \approx F \cdot \left( \min\{H, \ell_{\mathrm{AR},k}\} + g_k(H) + 1 \right).
\end{equation}
Here, $\ell_{\mathrm{AR},k}$ denotes the effective AR memory:
\begin{equation}
\label{eq:l_ar}
    \ell_{\mathrm{AR}, k} = \left\lceil \frac{\ln(1 / (1 - \epsilon))}{- \ln \rho_k} \right\rceil, \ \epsilon\in(0,1)
\end{equation}
where $\rho_k \in (0,1)$ is the spectral radius of the AR companion matrix. $g_k(H)$ reflects the resolved seasonal complexity:
\begin{equation}
    g_k(H) = 2 \sum_{j=1}^{J} w_{j,k} \cdot \min \left(1, \frac{H}{T_{j,k}^*} \right),
\end{equation}
\begin{equation}
    w_{j,k} = \frac{\sum_{f=1}^{F} A_{f,j,k}^{2}}{\sum_{f=1}^{F}\sum_{j=1}^{J} A_{f,j,k}^{2}}.
\end{equation}

This formulation yields a compact and information-preserving representation that enables a precise loss decomposition and supports optimal horizon analysis under federated, non-IID settings. Please refer to the Extended Version for more details.

\section{Loss Analysis}
\label{sec:loss_analysis}

Before we proceed to analyze how the look-back horizon $H$ affects forecasting performance in the federated setting, we explore decomposition of the prediction loss into two components: an intrinsic, irreducible term that captures the uncertainty of the data-generating process, and an approximation term that captures the difficulty of learning a mapping in an $H$-dimensional input space using a finite-capacity model trained on finite and heterogeneous data.

\subsection{Overall Loss Analysis}

Consider the forecasting task of predicting the next $S$ values from the previous $H$ observations. Under the intrinsic-space representation, this corresponds to learning a mapping $m: \mathcal{M}(H) \to \mathcal{M}(S)$. For a given client distribution and any measurable predictor $m$, the squared loss can be decomposed into
\begin{equation}
\label{eq:loss-decomposition}
    L(H,S;m) = L_{\mathrm{Bayes}}(H,S) + L_{\mathrm{approx}}(H,S;m)
\end{equation}
where:

\begin{itemize}
    \item \textbf{Bayesian loss} $L_{\mathrm{Bayes}}(H,S)$ is the irreducible error incurred even by an ideal predictor with full knowledge of the data distribution.

    \item \textbf{Approximation loss} $L_{\mathrm{approx}}(H,S;m)$ captures the additional error due to using a finite-capacity predictor $m$ trained on limited local data. 
\end{itemize}
The formal derivation of \eqref{eq:loss-decomposition} and the precise definitions of the two terms are given in the Extended Version.


We now formalize this intuition by establishing a precise decomposition of the prediction loss in the federated setting, showing how the Bayesian and approximation components arise directly from the client-specific data-generating distributions and the server-side evaluation protocol.

\begin{theorem}[Federated Loss Decomposition]
\label{thm:federated_loss_decomposition}

For each client $k \in \{1,\dots,K\}$, let $(U_k, V_k)$ denote its data-generating pair, where $U_k$ takes values in a measurable input space $\mathcal{M}(H)$ and $V_k$ in an output space $\mathcal{M}(S)$, both embedded in a real Hilbert space $(\mathcal{H},\|\cdot\|)$ with the associated Borel $\sigma$-algebras. 

Let $m_k^*(u) := \mathbb{E}[\ V_k \mid U_k = u\ ]$ be the client-specific Bayesian predictor, defined $P_{U_k}$–almost everywhere. For any measurable, square-integrable predictor $m : \mathcal{M}(H) \to \mathcal{M}(S)$, the server's global predictive loss is
\begin{equation}
    L(H,S;m) := \mathbb{E}_{k\sim\pi}\Big[ \mathbb{E} \big[\ \|\ V_k - m(U_k)\ \|^2 \big] \Big]
\end{equation}
where $\pi=(\pi_1,\dots,\pi_K)$ is any distribution over clients and the inner expectation is over $(U_k, V_k)$ under client $k$'s distribution.
Then the loss decomposes as: 
\begin{equation}
    L(H,S;m) = L_{\mathrm{Bayes}}(H,S) + L_{\mathrm{approx}}(H,S;m),
\end{equation}
where the federated Bayesian loss is
\begin{equation}
    L_{\mathrm{Bayes}}(H,S) := \mathbb{E}_{k\sim\pi}\Big[ \mathbb{E}\big[\ \|\ V_k - m_k^*(U_k)\, |^2 \big] \Big],
\end{equation}
and the federated approximation loss is
\begin{equation}
    L_{\mathrm{approx}}(H,S;m) := \mathbb{E}_{k\sim\pi}\Big[ \mathbb{E}\big[\ \|\ m_k^*(U_k) - m(U_k)\ \|^2 \big] \Big].
\end{equation}

In particular, the total loss separates into the expected irreducible (client-wise Bayes) component and the expected approximation error of the global predictor relative to each client’s Bayes-optimal rule.
Please refer to the Extended Version for the proof.

\end{theorem}

\paragraph{Server–client interpretation }

The global model $m$ is hosted on the central server and evaluated on clients sampled according to $k \sim \pi$. The term $L_{\mathrm{Bayes}}(H, S)$ captures the irreducible uncertainty within each client’s local
data-generating process, averaged over clients, while $L_{\mathrm{approx}}(H, S; m)$ measures the discrepancy between the global server model and the collection of Bayes-optimal per-client predictors $\{m_k^*\}_{k=1}^K$. Although both components are defined via client-side distributions, the total loss $L(H, S; m)$ represents the expected prediction error of the server’s global model.

In the remainder of this section, we investigate these two components respectively.

\subsection{Bayesian (Irreducible) Loss}

We first characterize the irreducible component of predictive loss for each client using the structure of the SDG.

\begin{theorem}[Client-wise Bayesian Loss]
\label{thm:client-bayes}

According to the SDG model in Equation~\eqref{eq:sdg} for client $k$: each feature is generated as an additive sum of (i) an autoregressive component, (ii) a seasonal component, (iii) a linear trend, and (iv) an innovation noise term, with the innovations independent across time and independent of the deterministic seasonal and trend components.
Then the client-wise Bayesian loss admits the exact decomposition:
\begin{equation}
\label{eq:client-bayes-final}
    L_{\mathrm{Bayes}}^{(k)}(H,S) = L_{\mathrm{AR}}^{(k)}(S) + L_{\mathrm{seas}}^{(k)}(H) + L_{\mathrm{trend}}^{(k)}(H)
\end{equation}
where each term is the contribution of the corresponding SDG component to
the conditional mean-squared error under a horizon-$H$ Bayesian predictor:
\begin{equation}
    L_{\mathrm{AR}}^{(k)}(S)
        := \mathbb{E}\big[
            \|\ Y_{\mathrm{AR},k}^{(S)}
              - \mathbb{E}[Y_{\mathrm{AR},k}^{(S)} \mid X_k^{(H)}] \ \|_2^2
        \big],
\end{equation}
\begin{equation}
        L_{\mathrm{seas}}^{(k)}(H)
        := \mathbb{E}\big[
            \|\ Y_{\mathrm{seas},k}^{(S)}
              - \mathbb{E}[Y_{\mathrm{seas},k}^{(S)} \mid X_k^{(H)}] \ \|_2^2
        \big],
\end{equation}
\begin{equation}
        L_{\mathrm{trend}}^{(k)}(H)
        := \mathbb{E}\big[
            \|\ Y_{\mathrm{trend},k}^{(S)}
              - \mathbb{E}[Y_{\mathrm{trend},k}^{(S)} \mid X_k^{(H)}] \ \|_2^2
        \big].
\end{equation}
Here $X_k^{(H)}$ and $Y_k^{(S)}$ denote the input window and $S$-step forecast block for client $k$, and
$Y_{\mathrm{AR},k}^{(S)}$, $Y_{\mathrm{seas},k}^{(S)}$, $Y_{\mathrm{trend},k}^{(S)}$ are the corresponding SDG components of the future block $Y_k^{(S)}$. Please refer to the Extended Version for the component-wise characterization and bounds.

\end{theorem}

\begin{remark}

For each client $k$, the Bayesian loss $L_{\mathrm{Bayes}}^{(k)}(H, S)$ decreases with the look-back horizon $H$ as longer histories improve identifiability of seasonal structure and (where present) trend components. The loss increases in the forecast horizon $S$, reflecting the accumulation of autoregressive innovations. Once the dominant seasonal cycles and the client's effective AR memory are covered, further increasing $H$ yields only negligible improvement: the Bayesian loss has reached its horizon-dependent saturation level.

\end{remark}

The irreducible uncertainty perceived by the server is the weighted combination of these client-level Bayesian losses.

\begin{lemma}[Server-level Bayesian Loss Aggregation]
\label{lemma:server-bayes}

Let $\pi=(\pi_1,\dots,\pi_K)$ be any probability distribution over the $K$ clients. The population-level Bayesian loss is
\begin{equation}
    L_{\mathrm{Bayes}}^{\mathrm{(server)}}(H,S) = \sum_{k=1}^K \pi_k\, L_{\mathrm{Bayes}}^{(k)}(H,S),
\end{equation}
a quantity determined by the client data-generating processes $(U_k, V_k)$ and independent of any global predictor. It aggregates the client-wise irreducible components (autoregressive variation, seasonal
residuals, and optional trend terms), each of which exhibits a distinct dependence on the horizon $H$ according to the client’s temporal dynamics.

\end{lemma}

\subsection{Approximation Loss}

We now analyze the approximation loss in the federated setting, where a global model $m$ is trained on a central server using client-local updates. This loss arises from the discrepancy between the global model and the Bayes-optimal predictor on each client.

\begin{theorem}[Client-wise Approximation Loss]
\label{thm:client-approx}

For client $k$, let $m_k^*(X)$ be the Bayesian predictor and $m$ be any learned predictor. The approximation loss at horizon $(H,S)$ is
\begin{equation}
    L_{\mathrm{approx}}^{(k)}(H,S; m) := \mathbb{E} \left[\ \|\, m(X) - m_k^*(X) \|_2^2 \ \right].
\end{equation}

Assume the Bayesian predictor $m_k^*$ is twice differentiable on the intrinsic representation space with bounded curvature, and that $m$ is a piecewise-affine model defined on the intrinsic manifold of dimension $d_{I,k}(H)$. Let $D_k$ denote the number of training windows on client $k$.

Then the approximation loss admits the intrinsic-dimension–dependent bound
\begin{equation}
\label{eq:approx-main-body}
    \begin{split}
        L_{\mathrm{approx}}^{(k)}(H,S; m)
            & \lesssim \Big( K_2^2\, d_{I,k}(H)^2 \Big)^{\frac{d_{I,k}(H)}{4 + d_{I,k}(H)}} \\
            & \quad + \Big( \frac{d_{I,k}(H)\, H}{D_k} \Big)^{\frac{4}{4 + d_{I,k}(H)}}
    \end{split}
\end{equation}
where $K_2$ is a curvature constant depending only on $m_k^*$. The first term reflects the geometric complexity of the intrinsic manifold, and the second term quantifies finite-sample limitations due to the effective sample size $D_k/H$. Full technical derivation is provided in the Extended Version.

\end{theorem}

The client-wise approximation losses aggregate  to form the global loss on the server:

\begin{lemma}[Server-level Approximation Loss Aggregation]
\label{lemma:server-approx}

Let $\pi = (\pi_1,\ldots,\pi_K)$ be the client-sampling distribution used by the server, with $\pi_k \ge 0$ and $\sum_k \pi_k = 1$. The global approximation loss under the server-side predictor $m$ is the weighted aggregation of the client-wise approximation losses:
\begin{equation}
    L_{\mathrm{approx}}^{(\mathrm{server})}(H,S; m) = \sum_{k=1}^{K} \pi_k \ L_{\mathrm{approx}}^{(k)}(H,S; m).
\end{equation}

\end{lemma}

\begin{remark}

Because the intrinsic dimension $d_{I,k}(H)$ typically increases with the look-back horizon $H$, and the number of effectively independent samples scales as $D_k/H$, both the curvature-driven bias term and the finite-sample variance term in $L_{\mathrm{approx}}^{(k)}(H, S; m)$ grow with $H$. Consequently, each client exhibits a horizon beyond which approximation error begins to dominate. The server-level approximation loss inherits this behavior via the mixture weights $\pi$, reflecting how rising intrinsic complexity and diminishing effective sample size jointly amplify the approximation error.

\end{remark}

\noindent
In this section, we decompose the forecasting error into its fundamental components and characterize how each behaves as a function of the look-back horizon $H$ and forecasting span $S$.   We begin by expressing the population prediction error as the sum of (i) the Bayes loss, which reflects irreducible uncertainty in the SDG, and (ii) the approximation loss, which arises from learning a nonlinear predictor from finite data.  This yields the client-wise decomposition with the global loss obtained by aggregation across clients.

The combined loss, therefore, exhibits a fundamental tradeoff: the Bayesian loss decreases and eventually plateaus, while the approximation loss increases with $H$.  Their interaction induces a unimodal structure
in the total loss $L^{(k)}(H, S)$ and yields a client-specific optimal look-back horizon that balances signal coverage with statistical efficiency. This analysis forms the theoretical basis for the optimal
horizon selection framework developed in the next section.

\section{Optimal Horizon $H$}\label{sec:optimal_h}

This section formalizes the choice of the look-back horizon $H$ as an explicit optimization over the total loss
\begin{equation}
    \label{eq:total-loss}
    L(H,S; m) = L_{\mathrm{Bayes}}(H,S) + L_{\mathrm{approx}}(H,S; m)
\end{equation}
where $L_{\mathrm{Bayes}}$ and $L_{\mathrm{approx}}$ are defined through the intrinsic-space formulation and federated loss decomposition in the previous sections. Since the trained model implicitly depends on the horizon, we slightly abuse notation and write $L(H, S)$ and $L_{\mathrm{approx}}(H, S)$ for $L(H, S; m)$ and  $L_{\mathrm{approx}}(H, S; m)$, suppressing the dependence on $m$ to avoid clutter.

We work client-wise (suppressing the client index when clear) and treat $H\in\mathbb{N}$, using forward differences
\begin{equation}
    \Delta f(H) := f(H{+}1) - f(H)
\end{equation}
to study how each loss component changes when one additional time step is added to the look-back window. Intuitively, the Bayesian loss should decrease as more history becomes available, while the approximation loss should increase due to higher intrinsic dimensionality and lower effective sample size. The remainder of this section quantifies this trade-off and characterizes the minimizer $H^*$.

\subsection{Bayesian–Approximation Loss Dynamics}

We study how the total client-wise prediction loss
\begin{equation}
    L^{(k)}(H) = L_{\mathrm{Bayes}}^{(k)}(H) + L_{\mathrm{approx}}^{(k)}(H; m)
\end{equation}
varies with the look-back horizon $H$.  We use the discrete forward difference $\Delta f(H):=f(H{+}1)-f(H)$ to analyze whether adding one additional step of history decreases or increases the loss.

\subsubsection{Bayesian Loss Behavior}

Under the SDG generative model (Eq.~\eqref{eq:sdg}), the irreducible loss decomposes into AR, seasonal, and (optionally zero) trend components. As $H$ increases, $\Delta L_{\mathrm{Bayes}}^{(k)}(H) \le 0$, and $\Delta L_{\mathrm{Bayes}}^{(k)}(H) \to 0$. \\

\noindent
\textit{Proof sketch.}
If $H_2 > H_1$, then $\sigma(X_{1:H_1}) \subseteq \sigma(X_{1:H_2})$, and thus $\operatorname{Var}(Y \mid X_{1:H_2}) \le \operatorname{Var}(Y \mid X_{1:H_1})$ almost surely; taking expectations yields $L_{\mathrm{Bayes}}^{(k)}(H_2) \le L_{\mathrm{Bayes}}^{(k)}(H_1)$. Under the SDG, the future depends on finite AR memory $\ell_{\mathrm{AR},k}$, finite seasonal periods $T_{f,j,k}$, and a linear trend; therefore, the conditional expectation becomes invariant once $H \ge H_k^\ast$, implying $\Delta L_{\mathrm{Bayes}}^{(k)}(H) \to 0$. The monotonic decrease, therefore, follows from the conditional-variance identity,
and the eventual plateau follows from the finite dependency structure of the SDG. Thus, the Bayesian loss enters a plateau where additional history yields negligible improvement.

\subsubsection{Approximation Loss Behavior}

The learned model must approximate the intrinsic mapping from past to future.
From the dimension-dependent bound (Theorem~\ref{thm:client-approx}),
\begin{equation}
    \label{eq:approx-bound}
    \begin{split}
        L_{\mathrm{approx}}^{(k)}(H;m)
        &\lesssim 
        \Big(K_2^2\, d_{I,k}(H)^2\Big)^{\tfrac{d_{I,k}(H)}{4+d_{I,k}(H)}}
        \\
        &\quad +
        \Big(\tfrac{d_{I,k}(H)\,H}{D_k}\Big)^{\tfrac{4}{4+d_{I,k}(H)}},
    \end{split}
\end{equation}
where $d_{I,k}(H)$ denotes the intrinsic dimension of the input window and
$D_k$ is the number of overlapping training windows on client $k$. \\

\noindent
\textit{Proof sketch.}
Both terms in \eqref{eq:approx-bound} worsen as $H$ grows. First, $d_{I,k}(H)$ is nondecreasing by construction, so the model must approximate a higher-dimensional function class; the dimension-dependent term 
$\big(K_2^2 d_{I,k}(H)^2\big)^{d_{I,k}(H)/(4+d_{I,k}(H))}$ therefore increases with $H$. Second, due to window overlap, the number of effectively independent samples scales as $D_k/H$; hence, the statistical term  
$\big(d_{I,k}(H)H/D_k\big)^{4/(4+d_{I,k}(H))}$   also increases with $H$ since the numerator grows and the effective sample size shrinks. Thus, for sufficiently large $H$, $\Delta L_{\mathrm{approx}}^{(k)}(H;m) > 0$, 
showing that the approximation cost eventually worsens once the horizon is
long enough.

Hence, the approximation loss exhibits the opposite trend of the Bayesian loss: increasing window size ultimately leads to higher approximation error.

\subsubsection{Smallest sufficient horizon}

Now we define a key concept, the smallest sufficient horizon, which serves as the optimal look-back horizon that minimizes the forecasting loss.

Formally, for any tolerance $\delta>0$, define the smallest sufficient horizon as
\begin{equation}
    H_k^*(\delta) := \min\{H : |\Delta L_{\mathrm{Bayes}}^{(k)}(H)| \le \delta\},
\end{equation}
at which the Bayesian loss has effectively saturated: further historical context improves the irreducible loss by at most $\delta$. Together, these monotonicity properties imply a unimodal structure for the
total loss.

\begin{theorem}[Unimodality and Optimal Horizon]
\label{thm:unimodal-2page}

If for a given $\delta>0$ the Bayesian loss satisfies $\Delta L_{\mathrm{Bayes}}^{(k)}(H) \le -\delta$ for all $H < H_k^*(\delta)$, and the approximation loss satisfies $\Delta L_{\mathrm{approx}}^{(k)}(H;m) \ge \delta$ for all $H \ge H_k^*(\delta)$, then the combined loss obeys that $L^{(k)}(H)$ decreases on $[1, H_k^*(\delta)]$, and $L^{(k)}(H)$ increases on $[H_k^*(\delta),\infty)$. 

Consequently, $H_k^*(\delta) \in \arg\min_{H\in\mathbb{N}} L^{(k)}(H)$ with uniqueness up to integer ties.

\end{theorem}

\begin{proof}

From the Bayesian loss analysis, increasing $H$ reduces seasonal/phase ambiguity and uncovers AR structure, but only up to a finite coverage horizon. Hence, there exists $H_0$ such that
\begin{equation}
    \Delta L_{\mathrm{Bayes}}(H,S) < 0 \quad (H < H_0),
\end{equation}
while for any $\delta>0$ we can choose $H_0$ large enough so that
\begin{equation}
    \Delta L_{\mathrm{Bayes}}(H,S) \ge -\delta \quad (H \ge H_0).
\end{equation}

For the approximation term, the curvature–variance bound on the intrinsic manifold shows that the error grows with both the intrinsic dimension $d_I(H)$ and the factor $H/D$ coming from the effective sample size per window ($\propto D/(HN)$). Since $d_I(H)$ is non-decreasing and eventually saturated, while $H/D$ grows linearly, there exists $\eta>0$, independent of $H$, such that
\begin{equation}
    \Delta L_{\mathrm{approx}}(H,S) \ge \eta \quad (H \ge H_0).
\end{equation}

Fix any $\delta \in (0,\eta)$ and define $H^*(\delta)$ as the smallest $H \ge H_0$ with $\Delta L_{\mathrm{Bayes}}(H,S) \ge -\delta$. Then for $H < H^*(\delta)$, we have $\Delta L_{\mathrm{Bayes}}(H,S) < -\delta$ and $\Delta L_{\mathrm{approx}}(H,S) \ge 0$, so $ \Delta L(H,S) = \Delta L_{\mathrm{Bayes}}(H,S) + \Delta L_{\mathrm{approx}}(H,S) < -\delta < 0$, and $L(H,S)$ is strictly decreasing. For $H \ge H^*(\delta)$, we have $\Delta L_{\mathrm{Bayes}}(H,S) \ge -\delta$ and $\Delta L_{\mathrm{approx}}(H,S) \ge \eta$, hence $\Delta L(H,S) \ \ge \ -\delta + \eta \ > \ 0$, so $L(H,S)$ is strictly increasing.

Thus $L(H, S)$ decreases up to $H^*(\delta)$ and increases thereafter, so it is unimodal in $H$ and attains its unique minimum at $H^*(\delta)$ (up to trivial ties), as claimed.

\end{proof}

\noindent
Hence, before $H_k^*(\delta)$, the reduction in irreducible error outweighs the increase in approximation error; afterwards, the opposite holds. The total loss thus has a single optimal basin, and the smallest sufficient horizon attains the minimum.

\subsubsection{Seasonal Coverage and Horizon Selection}

The tolerance $\delta$ can be linked to an interpretable signal structure via seasonal coverage. Let $A_k^2 = \sum_{f,j}A_{f,j,k}^2$ denote the total seasonal energy of client $k$, and define the $\tau$-coverage horizon $T_k^{(\tau)}$ as the smallest $H$ for which the unresolved seasonal energy beyond $H$ obeys:
\begin{equation}
    \sum_{f,j : T_{f,j,k} > H} A_{f,j,k}^2 \ \le\ (1-\tau)\ A_k^2,
\end{equation}
assuming the residual seasonal loss satisfies
$L_{\mathrm{seas}}^{(k)}(H)\le A_k^2r(H,T)$ with $r(\cdot,T)$ decreasing in
$H$.

\begin{corollary}[Coverage–Tolerance Mapping]
\label{cor:coverage-2page}

If a coverage level $\tau$ is chosen so that $(1-\tau)A_k^2 \le \delta$, then every $H \ge T_k^{(\tau)}$ satisfies $|\Delta L_{\mathrm{Bayes}}^{(k)}(H)|\le\delta$. Thus, the optimal horizon is given by
\begin{equation}
    H_k^*(\delta) = \max\{\ell_{\mathrm{AR},k},\ T_k^{(\tau)}\}.
\end{equation}

\end{corollary}

This provides a direct way to set the horizon using interpretable signal parameters: choose a desired seasonal coverage $\tau$, infer the corresponding $\delta$, and compute $H_k^*(\delta)$.

\subsubsection{Federated Horizon Aggregation}

In federated learning, the server must choose a single global horizon $H_{\mathrm{server}}$ despite heterogeneous client optima $\{H_k^*(\delta)\}$. Because extreme clients (e.g., very large horizons) can substantially reduce effective sample sizes for all participants, robustness is crucial. Let $w_k \propto n_k$ be data-proportional weights normalized so that $\sum_k w_k = 1$.

\subsubsection{Robust Federated Horizon}

The global horizon can be defined via the weighted trimmed mean:
\begin{equation}
    H_{\mathrm{server}}^* = \operatorname{TrimMean}_{\alpha}\big(\{H_k^*(\delta)\}_{k=1}^K ; \{w_k\}_{k=1}^K\big)
\end{equation}
which discards an $\alpha$-fraction of the smallest and largest client-specific horizons (by weight) and averages the remainder.

This estimator is equivalent to minimizing a convex Huber-type aggregation objective and yields a horizon that balances most clients while avoiding inflation by a small number of extreme ones.

\section{Discussion and Conclusion}
\label{sec:conclusion}

\subsection{Limitation and Discussion}

This work proposes a principled framework for federated time-series forecasting under non-IID conditions, grounded in a structured synthetic data generator (SDG) and an intrinsic space formulation. The framework enables a precise decomposition of forecasting error and leads to a provably optimal look-back horizon. To make the analysis tractable and the theoretical guarantees possible, several assumptions are made that define the scope of applicability.

The SDG models additive components, e.g., trend, autoregressive memory, and seasonality, with Gaussian innovations. While this structure captures core dynamics observed in real-world data, it does not account for regime switches, nonlinear seasonal patterns, or cross-feature interactions beyond what is implicitly represented through PCA. The analysis assumes local stationarity and a stable autoregressive structure, which may be challenged in long-memory or near-unit-root settings. Additionally, estimating global covariance in a federated context requires secure or privacy-aware aggregation, and our sample efficiency analysis treats overlapping windows as approximately independent, which may overstate the effective sample size under certain data regimes.

These assumptions are common in theoretical work and are intentionally chosen to isolate the role of horizon length and data heterogeneity. Importantly, they enable the first provable characterization of optimal look-back windows in federated forecasting, providing a foundation for future extensions that relax these constraints.

\subsection{Conclusion}

This paper introduces a principled framework for horizon selection in federated time series forecasting under non-IID conditions, grounded in a synthetic data generator (SDG) that models key temporal structures, trend, autoregressive memory, and seasonality, along with client-specific heterogeneity. By embedding time-series windows into a geometry-preserving intrinsic space, we enable a precise decomposition of forecasting loss into irreducible Bayesian error and model-dependent approximation error, each tied to the underlying statistical structure and data distribution. Our analysis reveals a fundamental trade-off: while the Bayesian loss decreases with horizon length as more temporal structure becomes identifiable, the approximation loss increases due to growing intrinsic dimension and reduced sample efficiency. This yields a provable result that the total loss is minimized at the smallest sufficient horizon $H^*$, where additional history no longer improves identifiability but exacerbates overfitting. Furthermore, we propose a robust aggregation strategy to identify a global horizon across clients. Together, these contributions establish the first theoretically grounded criterion for adaptive horizon selection in federated settings, offering practical guidance for model design, deployment, and benchmarking in decentralized, heterogeneous environments.

\section{Acknowledgments}

This work is partly supported by the Australian Research Council Linkage Project (Grant No. LP220200893).

\bibliography{aaai2026}

\section{Appendix}

\subsection{Appendix A: SDG Experimental Validation}

We validate the SDG by decomposing a real series into interpretable components (linear trend, sinusoidal seasonality, and AR($p$) residuals), then re-synthesizing a signal with the fitted pieces and comparing it back to the original across marginal, temporal, and spectral views.

\subsubsection{Dataset}

The temperature time series used in this study was extracted from the Max Planck Institute for Biogeochemistry’s weather portal (\url{https://www.bgc-jena.mpg.de/wetter/}), which provides 10-minute resolution meteorological data recorded in Jena, Germany, throughout the year 2020. The full dataset includes 21 weather indicators (e.g., air temperature, humidity, wind speed), but only the air temperature series was used in this experiment. A temperature time series at 10-minute resolution ($N = 52{,}696$) is parsed as $y_t$, $t = 1, \dots, N$.

\subsubsection{Hyperparameters}

Trend is estimated via OLS ($y_t \approx \beta t + c$); dominant periods are selected from the periodogram (daily peak $T_1=144$ plus an ultra–low-frequency component tied to record length); detrended residuals are fit with AR models, with order chosen by AIC/BIC (adopt $p=30$) and Gaussian innovation scale $\sigma$ taken from the residual standard deviation.

\subsubsection{Metrics}

(1) Mean/variance gaps to check marginal alignment; (2) ACF $L^2$ gap up to 30 lags for short-range dependence; (3) normalized PSD $L^2$ gap for spectral structure; (4) two-sample KS statistic for distributional shape; (5) window-level (50-step) real-vs-synthetic discrimination accuracy of a random-forest classifier to probe downstream indistinguishability.

\subsubsection{Result}

The SDG reproduces the original series closely: $\Delta\mu\approx 3.6\!\times\!10^{-3}$, ACF $L^2\!\approx\!3.7\!\times\!10^{-6}$, normalized PSD $L^2\!\approx\!8.2\!\times\!10^{-3}$, KS $=0.042$ (large-$N$ significance), and a RF accuracy of $0.892$ on 50-step windows, indicating strong agreement in dependence/spectral structure with minor residual cues useful for further tuning.

\subsection{Appendix B: Intrinsic Space Construction}

In this section, we provide a step-by-step transformation to convert a non-IID time series data of length $L$, which can be well described by the SDG, into a vector $z$ in an intrinsic space $M(L)$.

Let $O(H) \subset \mathbb{R}^{FH}$ be the space of normalized windows where $H$ is the window length for one sample of a time series data. We construct $\Phi_H: O(H) \rightarrow M(H)$ satisfying the following assumptions:

\begin{assumption}[Compact image with uniform radius control]\label{asmp:bounded-radius}
For each horizon $H\in\mathbb{N}$ there exists a finite radius $R_H$ such that
\begin{equation}
    \sup_{x\in O(H)}\|\Phi_H(x)\|_2 \;\le\; R_H,
\end{equation}

so that $M(H):=\Phi_H(O(H))\subseteq \overline{B}_{R_H}(0)\subset \mathbb{R}^{d_I(H)}$. $\overline{B}_{R_H}(0)$ is the closed Euclidean ball of the radius $R_H$: 

\begin{equation}
    \overline{B}_{R_H}(0) := \left\{ y \in \mathbb{R}^{d_{I(H)}} : \|y\|_2 \leq R_H \right\}
\end{equation}
\end{assumption}

\begin{remark}
This keeps all intrinsic representations within a controlled region. It ensures basic quantities (norms, averages, covariances) are well-behaved and comparable across horizons.
\end{remark}

\begin{assumption}[Bi-Lipschitz embedding with stable inverse]\label{asmp:bi-lipschitz}
For each horizon $H$ there exist constants $\alpha_H>0$ and $\beta_H<\infty$ such that for all $x,y\in O(H)$
\begin{equation}
\alpha_H\,\|x-y\|_2 \;\le\; \|\Phi_H(x)-\Phi_H(y)\|_2 \;\le\; \beta_H\,\|x-y\|_2.
\end{equation}
Consequently, $\Phi_H$ is injective and continuous, and its inverse on $M(H)$ is Lipschitz with constant at most $1/\alpha_H$. Define the distortion $\kappa_H:=\beta_H/\alpha_H$.
\end{assumption}

\begin{remark}
This preserves distances up to a fixed factor when moving between window space and intrinsic space. It lets us transfer geometric and statistical statements between the two spaces with controlled distortion.
\end{remark}

\begin{assumption}[Intrinsic dimension: regularity, monotonicity, and saturation]\label{asmp:intrinsic-dim}
For each horizon $H \in \mathbb{N}$, the image $M(H) := \Phi_H(O(H))$ is assumed to be a connected, continuously differentiable ($C^1$) submanifold of $\mathbb{R}^{d_I(H)}$ with intrinsic dimension equal to the embedding dimension, i.e., $\dim M(H) = d_I(H)$. Moreover, the intrinsic dimension is non-decreasing with horizon, satisfying $d_I(H+1) \ge d_I(H)$ for all $H$. Finally, there exists a finite \emph{saturation horizon} $H_{\mathrm{id}}$ such that for all $H \ge H_{\mathrm{id}}$, the intrinsic dimension stabilizes: $d_I(H) = d_I(H_{\mathrm{id}})$.
\end{assumption}

\begin{remark}
This assumption encodes a structural view of how complexity in the representation evolves with increasing horizon: initially, as $H$ grows, the map $\Phi_H$ incorporates more information and thus the intrinsic dimensionality of its image increases. Once a sufficient horizon is reached, where it captures all essential long-term structure, the complexity no longer grows. The saturation point $H_{\mathrm{id}}$ marks the minimal horizon beyond which the representation is informationally complete.
\end{remark}

\begin{assumption}[Inter-horizon compatibility via a stable linear map]\label{asmp:inter-horizon}
For any $H_1 \le H_2$, there exists a linear map $P[H_2,H_1]: \mathbb{R}^{d_I(H_2)} \to \mathbb{R}^{d_I(H_1)}$ that preserves distances on $M(H_2)$ up to a uniform distortion factor. Specifically, there exists a constant $C_{\mathrm{iso}} \ge 1$, independent of $H_1$ and $H_2$, such that for all $z_1, z_2 \in M(H_2)$,

\begin{equation}
\begin{split}
    C_{\mathrm{iso}}^{-1}\,\|z_1-z_2\|_2 & \;\le\; \|P[H_2,H_1](z_1)-P[H_2,H_1](z_2)\|_2 \\
& \;\le\; C_{\mathrm{iso}}\,\|z_1-z_2\|_2.
\end{split}
\end{equation}
\end{assumption}

\begin{remark}
This gives a simple, stable way to relate representations at different horizons. It supports comparing statistics and constructions across $H$ using a single linear operator.
\end{remark}

\begin{assumption}[Truncation behaves like projection up to a small error]\label{asmp:truncation-projection}
Let $t_p[H_2, H_1] : O(H_2) \to O(H_1)$ denote the truncation map that discards the last $H_2 - H_1$ time steps. Assume that the representation map $\Phi_{H_2} : O(H_2) \to M(H_2)$ is a bijection onto its image. Then, define the induced intrinsic truncation map:

\begin{equation}
    \mathcal{T}_{H_2 \to H_1} := \Phi_{H_1} \circ t_p[H_2, H_1] \circ \Phi_{H_2}^{-1} : M(H_2) \to \mathbb{R}^{d_I(H_1)}
\end{equation}

There exist constants $c_{\mathrm{err}} > 0$ and $\gamma \in [1, 2]$ such that for all $z \in M(H_2)$,

\begin{equation}
\left\| \mathcal{T}_{H_2 \to H_1}(z) - P[H_2, H_1]\,z \right\|_2 \;\le\; c_{\mathrm{err}}\,H_1^{-\gamma}
\end{equation}

where $P[H_2, H_1]: \mathbb{R}^{d_I(H_2)} \to \mathbb{R}^{d_I(H_1)}$ is the stable linear projection map from Assumption~\ref{asmp:inter-horizon}.
\end{assumption}

\begin{remark}
This assumption states that truncating the input sequence and then applying the lower-horizon representation map is approximately equivalent to linearly projecting the higher-horizon representation. The approximation error decays with horizon length, reflecting the idea that the truncation becomes less destructive as $H_1$ grows.
\end{remark}

\begin{assumption}[Power-law decay of intrinsic covariance eigenvalues]\label{asmp:zipf}
Let $\Sigma_H$ be the intrinsic covariance on $M(H)$ with eigenvalues $\lambda_1(H)\ge\lambda_2(H)\ge\cdots$. There exist constants $\alpha_Z\in(1,2]$ and $C_Z>0$ such that for all $i\ge 1$ and all $H$
\begin{equation}
\lambda_i(H) \;\le\; C_Z\,i^{-\alpha_Z}.
\end{equation}
Equivalently, the tail energy beyond the top $d$ components decays on the order of $d^{\,1-\alpha_Z}$, uniformly in $H$.
\end{assumption}

\begin{remark}
This says most variance sits in a few leading directions, making low-dimensional summaries effective. It supports later bounds that depend on how quickly energy concentrates in the intrinsic representation.
\end{remark}

Now we propose the step-by-step transformation converting a non-IID time series data described by the SDG into an intrinsic space:

\begin{enumerate}
    \item \textbf{Client-wise Normalization:} To reduce the impact of feature skew caused by the aﬀine skew, we apply the client-wise local normalization, allowing data from different clients to be aligned into a common representation space. For each client $k$ and feature $f$, compute:
    \begin{equation}
        \mu_{f,k} = \frac{1}{H} \sum_{i=1}^H x_{f,i,k},
    \end{equation}
    \begin{equation}
        \sigma_{f,k} = \sqrt{\frac{1}{H} \sum_{i=1}^H (x_{f,i,k} - \mu_{f,k})^2}
    \end{equation}
    and normalize:
    \begin{equation}
        \hat{x}_{f,i,k} = \frac{x_{f,i,k} - \mu_{f,k}}{\sigma_{f,k}} \quad \text{for all } f, i
    \end{equation}

    \item \textbf{Flatten Windows into Vectors:} To analyze the global structure, we flatten the windows into vectors, which is suitable for further PCA analysis. Each normalized window $\hat{X}_{f,t,k}^{(H)} \in \mathbb{R}^{F \cdot H}$ is flattened (e.g., row-wise) into:
    \begin{equation}
        \hat{x}_{t,k}^{\text{flat}} = \text{vec}(\hat{X}_{t,k}) \in \mathbb{R}^{F \cdot H}
    \end{equation}

    \item \textbf{Global Covariance and Eigendecomposition:} To identify principal directions of variation in the normalized time series data, we compute the empirical covariance matrix of the flattened windows. Let $X \in \mathbb{R}^{N \times (F \cdot H)}$ be the matrix whose rows are the centered, flattened vectors $\hat{x}^{\text{flat}}_{t,k}$. We compute:
    \begin{equation}
        \Sigma = \frac{1}{N} X^\top X
    \end{equation}
    and perform eigendecomposition:
    \begin{equation}
        \Sigma = U \Lambda U^\top
    \end{equation}
    The matrix $U$ contains the principal directions of variation. When the data lies near a low-dimensional linear (or locally linear) submanifold, this eigenspace approximates the intrinsic structure. The eigenvalues in $\Lambda$ reflect the energy captured by each direction.

    \item \textbf{Estimate Intrinsic Dimension $d_{I,k}(H)$:} The intrinsic dimension $d_I(H)$ reflects the number of effective degrees of freedom in a time series window of size $H$ for $F$ features. Since the SDG is composed of a small number of structured, decoupled components (including AR memory, seasonality, and trend), the observed data must lie near a low-dimensional manifold whose dimension is approximately the sum of degrees of freedom required to encode each component. Hence, we estimate the intrinsic dimension $d_I(H)$ based on the modeled components of the SDG:

    \paragraph{AR memory.} Let $\rho_k\!\in\!(0,1)$ be the spectral radius of the AR companion matrix. Define the effective memory length
    \begin{equation}
        \label{eq:l_ar}
        \ell_{\mathrm{AR},k} = \tau_{e,k} = \left\lceil \frac{\ln(1/(1-\epsilon))}{-\ln \rho_k} \right\rceil, \ \epsilon\in(0,1)
    \end{equation}
    The e-folding time for the impulse response to decay to a fraction $\epsilon$. (Empirically, the ACF $1/e$ crossing is a similar estimate.)

    \paragraph{Seasonality (amplitude-weighted).} Each sinusoid contributes two degrees of freedom once \emph{one full cycle} is observed. Let
    \begin{equation}
        g_k(H) = 2\sum_{j=1}^{J} w_{j,k}\,\min\!\Big(1,\frac{H}{T_{j,k}^{*}}\Big)
    \end{equation}
    \begin{equation}
        w_{j,k} = \frac{\sum_{f=1}^{F} A_{f,j,k}^{2}}{\sum_{f=1}^{F}\sum_{j=1}^{J} A_{f,j,k}^{2}}
    \end{equation}
    where $T_{j,k}^{*}$ is the (feature-aggregated) period for component $j$ on client $k$.

    \paragraph{Trend and total intrinsic dimension.} Add $+1$ for linear trend. The resulting intrinsic dimension is
    \begin{equation}
        \begin{split}
            d_{I,k}(H)
            &\approx F\cdot\Big(\min\{H,\ell_{\mathrm{AR},k}\}+g_k(H)+1\Big)\,,
        \end{split}
        \label{eq:dim_intrinsic}
    \end{equation}
    which is smooth, monotone in $H$, and saturates once $H\!\ge\!\ell_{\mathrm{AR},k}$ and $H\!\ge\!\max\{T_{j,k}^{*}: w_{j,k}\text{ significant}\}$.
    
    The intrinsic dimension $d_I(H)$ can also be estimated empirically to verify that the theoretical $d_I(H)$ agrees with the actual spectrum and provides a quick plug-in value:
    \begin{equation}
        \begin{split}
            \frac{\sum_{i=1}^{d_I} \lambda_i}{\sum_{j=1}^{F \cdot H} \lambda_j} \geq \eta \quad \text{(e.g., } \eta = 0.99\text{)}
        \end{split}
    \end{equation}
    This step provides a principled way to compute $d_I$ from the generative process structure, ensuring that the low-dimensional representation retains core dynamics.

    \item \textbf{Step 5: Project to Intrinsic Space}
    Let $U_{d_I} \in \mathbb{R}^{F \cdot H \times d_I}$ be the matrix of top eigenvectors. We project the flattened time series data into an intrinsic space:
    \begin{equation}
        \begin{split}
            z_{t,k} &= \Phi_H(x_{t,k}) \\
            &= U_{d_I}^\top \cdot \hat{x}_{t,k}^{\text{flat}} \in \mathbb{R}^{d_I} \\
            &= U_{d_I}^\top \cdot \text{vec}\left( \frac{X_{t,k} - \mu_k}{\sigma_k} \right)
        \end{split}
    \end{equation}
\end{enumerate}

Now we have shown the detailed steps of transforming typical real-world non-IID time series data described by the SDG into an intrinsic space. In words, the intrinsic space formulation offers a unified geometric and statistical foundation for representing time series data in federated settings.

\subsection{Appendix C: Loss Analysis}

In this section, we include supplementary materials for the Section Loss Analysis, starting with the formal definition of the Bayesian loss and the Approximation loss.

\begin{definition}[Bayesian (irreducible) loss]
\label{def:bayes-loss}

Fix a horizon pair $(H,S)\in\mathbb{N}^2$. Let $(U, V)$ be square-integrable random elements with $U \in \mathcal{M}(H)$, $V \in \mathcal{M}(S)$, defined on a common probability space and induced by the data-generating process (e.g., the SDG).

Since $V \in L^2$ and $U$ is measurable, the conditional expectation $\mathbb{E}[V \mid U]$ exists in $L^2$ and admits a measurable version defined almost everywhere. We denote any such version by the \emph{Bayes predictor}:
\begin{equation}
    m^*(u) := \mathbb{E}[\ V \mid U = u \ ], \quad \text{a.e. } u.   
\end{equation}

The \emph{Bayesian (irreducible) loss} at $(H,S)$ is
\begin{equation}
    L_{\mathrm{Bayes}}(H,S) := \mathbb{E} \left[\ \|\ V - m^*(U)\ \|_2^2\ \right].
\end{equation}

Equivalently, $L_{\mathrm{Bayes}}(H,S)$ is the minimum achievable risk
under squared loss over all measurable predictors
$m:\mathcal{M}(H)\!\to\!\mathcal{M}(S)$:
\begin{equation}
    L_{\mathrm{Bayes}}(H,S) = \inf_{m} \mathbb{E} \left[\ \|V - m(U)\|_2^2\ \right].
\end{equation}
    
For client $k$, define $L^{(k)}_{\mathrm{Bayes}}(H,S)$ analogously with
$(U,V)$ replaced by $(U_k,V_k)$.
\end{definition}

\begin{definition}[Approximation (excess) loss]
\label{def:approx-loss}

Let $m:\mathcal{M}(H)\to\mathcal{M}(S)$ be any square-integrable
predictor. The \emph{approximation loss} of $m$ at $(H,S)$ is the expected squared
distance to the Bayes predictor:
\begin{equation}
    L_{\mathrm{approx}}(H,S; m) := \mathbb{E} \left[\ \|\ m(U) - m^*(U)\ \|_2^2\ \right]
\end{equation}
where $m^*(U) := \mathbb{E}[V \mid U]$ is the Bayes predictor from
Definition~\ref{def:bayes-loss}. This quantity measures the excess risk of $m$ over the Bayes-optimal predictor:
\begin{equation}
    \mathbb{E}\!\left[\|V - m(U)\|_2^2\right] = L_{\mathrm{Bayes}}(H,S) + L_{\mathrm{approx}}(H,S; m)
\end{equation}

For client $k$, the client-specific approximation loss is
\begin{equation}
    L^{(k)}_{\mathrm{approx}}(H,S; m) := \mathbb{E}\!\left[\,\|\,m(U_k) - m_k^*(U_k)\,\|_2^2\,\right]
\end{equation}
where $m_k^*(U_k) := \mathbb{E}[V_k \mid U_k]$ is the client-wise Bayes predictor.

\end{definition}

The following proof will show that the total loss can be decomposed into the Bayesian loss and the approximation loss.

\begin{proof}[Proof of Total Loss Decomposition]

Let $X$ be the model input and $Y$ be the optimal output. By the nature of time series data, with \(Y \in L^2\), and let \(m: \mathbb{R}^d \to \mathbb{R}^p\) be any (possibly learned) predictor such that \(m(X) \in L^2\). Define the Bayes predictor as $m^*(X) := \mathbb{E}[Y \mid X]$, which is the \emph{orthogonal projection} of \(Y\) onto the closed subspace of \(L^2\) consisting of functions of \(X\). In Hilbert space terms, this means $Y = m^*(X) + R$, where the residual \(R := Y - m^*(X)\) satisfies:

\begin{equation}
    \mathbb{E}[\langle R,\ f(X) \rangle] = 0 \quad \text{for all } f(X) \in L^2.
\end{equation}

In particular, for any predictor \(m(X)\), the difference \(m^*(X) - m(X)\) is itself a function of \(X\), so we have:

\begin{equation}
    \mathbb{E}\left[\langle Y - m^*(X),\ m^*(X) - m(X) \rangle\right] = 0.
\end{equation}

Now, using the identity:

\begin{equation}
    \|Y - m(X)\|^2 = \|Y - m^*(X) + m^*(X) - m(X)\|^2.
\end{equation}

We expand the squared norm:

\begin{equation}
    \begin{split}
        \|Y - m(X)\|^2 &= \|Y - m^*(X)\|^2 + \|m^*(X) - m(X)\|^2 \\
        &\quad + 2 \langle Y - m^*(X),\ m^*(X) - m(X) \rangle.
    \end{split}
\end{equation}

Taking expectations on both sides and applying the orthogonality condition above, we obtain:

\begin{equation}
    \begin{split}
        \mathbb{E} & [\|Y - m(X)\|^2] \\
        &= \mathbb{E}[\|Y - m^*(X)\|^2] + \mathbb{E}[\|m^*(X) - m(X)\|^2] \\
        &\quad + 2 \ \mathbb{E}\left[\langle Y - m^*(X),\ m^*(X) - m(X) \rangle\right] \\
        &= \mathbb{E}[\|Y - m^*(X)\|^2] + \mathbb{E}[\|m^*(X) - m(X)\|^2].
    \end{split}
\end{equation}

Thus, the total predictive loss decomposes into:

\begin{equation}
    L = L_{\mathrm{Bayes}} + L_{\mathrm{approx}},
\end{equation}

where

\begin{equation}
    L_{\mathrm{Bayes}} := \mathbb{E}[\|Y - m^*(X)\|^2],
\end{equation}
\begin{equation}
    L_{\mathrm{approx}} := \mathbb{E}[\|m^*(X) - m(X)\|^2].
\end{equation}

\end{proof}

The following proof will show the representation of the Bayesian loss under the setting of the SDG and the intrinsic space.

\begin{proof}[Proof of Bayesian Loss Decomposition (Theorem~\ref{thm:client-bayes})]
Let $L^{(k)}_{\mathrm{Bayes}}(H, S)$ denote the irreducible forecasting loss for client $k$ when predicting $S$ future steps from a look-back window of length $H$. Under the SDG model (Equation~\ref{eq:sdg}), the time series consists of three additive and statistically independent components:
\begin{enumerate}
    \item An autoregressive (AR) component with order $p$ and client-specific coefficients $\phi_{k,i}$,
    \item A sum of deterministic sinusoidal (seasonal) signals,
    \item A deterministic linear trend (possibly optional).
\end{enumerate}

Because these components are independent, their contribution to the total irreducible error (i.e., Bayes loss) is additive:

\begin{equation}
    L^{(k)}_{\mathrm{Bayes}}(H,S) = L^{(k)}_{\mathrm{AR},\mathrm{Bayes}}(S) + L^{(k)}_{\mathrm{seas},\mathrm{Bayes}}(H) + L^{(k)}_{\mathrm{trend}}.
\end{equation}

We analyze each term individually: \\
 
\noindent \textbf{(1) Autoregressive (AR) Component.}  
Let $\Phi_k(z) = 1 - \sum_{i=1}^p \phi_{k,i} z^i$ denote the AR characteristic polynomial, and let $\psi_{k,s}$ be the impulse response of the equivalent MA($\infty$) process resulting from AR inversion:
\begin{equation}
    \psi_{k,0} = 1, \quad \psi_{k,s} = \sum_{i=1}^{p} \phi_{k,i} \psi_{k,s-i} \quad \text{for } s > 0.
\end{equation}

Once the AR coefficients $\phi_{k,i}$ are identified (i.e., $H \ge p$), the residual uncertainty in future values arises purely from $S$ future innovations, each filtered through the MA coefficients. Thus, the AR contribution to the Bayes loss is:

\begin{equation}
   L^{(k)}_{\mathrm{AR},\mathrm{Bayes}}(S) = \sum_{f=1}^{F} \sigma_{f,k}^2 \sum_{s=0}^{S-1} \psi_{k,s}^2, 
\end{equation}

where $\sigma_{f,k}^2$ is the innovation variance for feature $f$ on client $k$.

This term grows with $S$, but remains independent of $H$ once the AR parameters are recovered. Using geometric decay of the impulse response ($\psi_{k,s} \lesssim \rho_k^s$), we obtain the bound:

\begin{equation}
    L^{(k)}_{\mathrm{AR},\mathrm{Bayes}}(S) \le \sum_{f=1}^{F} \sigma_{f,k}^2 \cdot \frac{1 - \rho_k^{2S}}{1 - \rho_k^2},
\end{equation}

where $\rho_k$ is the spectral radius of the AR companion matrix.\\
 
\noindent \textbf{(2) Seasonal Component.}  
Each seasonal term is modeled as a sinusoid:

\begin{equation}
    A_{f,j,k} \cdot \sin\left( \frac{2\pi t}{T_{f,j,k}} + \theta_{f,j,k} \right).
\end{equation}

With infinite context, these deterministic components can be predicted perfectly. However, with a finite horizon $H$, there is residual uncertainty due to phase and frequency aliasing (i.e., spectral leakage). The residual energy for each sinusoid decays as the window covers more of its period:

\begin{equation}
    L^{(k)}_{\mathrm{seas},\mathrm{Bayes}}(H) \le \sum_{f=1}^{F} \sum_{j=1}^{J} A_{f,j,k}^2 \cdot r(H, T_{f,j,k}),
\end{equation}

where the decay function satisfies $r(H, T) \le c \cdot \min\left(1, (T/H)^{\gamma} \right)$ for some $\gamma \in [1,2]$, depending on the estimator's spectral resolution. As $H$ increases, this term decreases monotonically and saturates when $H \ge T_{f,j,k}$ for all dominant components. \\
 
\noindent \textbf{(3) Trend Component.}  
The linear trend $\beta_{f,k} t$ is fully deterministic and perfectly predictable when modeled. Thus, if the trend is included explicitly in the predictor class, it contributes no irreducible loss:

\begin{equation}
   L^{(k)}_{\mathrm{trend}} = 0. 
\end{equation}

Otherwise, if left unmodeled, it contributes a fixed additive term proportional to the variance of the time index $t$ over the prediction window. \\
 
\noindent \textbf{Conclusion.}  
Summing the three components yields the upper bound:

\begin{equation}
    \begin{split}
        L^{(k)}_{\mathrm{Bayes}}(H,S)
        &\le \sum_{f=1}^{F} \sigma_{f,k}^2 \cdot \frac{1 - \rho_k^{2S}}{1 - \rho_k^2} \\
        &\quad + \sum_{f=1}^{F} \sum_{j=1}^{J} A_{f,j,k}^2 \cdot r(H, T_{f,j,k}) \\
        &\quad + L^{(k)}_{\mathrm{trend}}.
    \end{split}
\end{equation}

This confirms the decomposition in Theorem~\ref{thm:client-bayes}, where each term reflects a distinct structural contribution from the SDG.

\end{proof}

The following proof demonstrates the decomposition of the approximation loss into the curvature and variance components in the federated scenario.

\begin{proof}[Proof of Approximation Loss Decomposition and Bound]

Fix a client $k$. Recall that the approximation loss is
\begin{equation}
    L_{\mathrm{approx}}^{(k)}(H,S;m) := \mathbb{E}\!\left[\|\,m_k^\star(U_k)-m(U_k)\|_2^2\right]
\end{equation}
where $m_k^\star := \mathbb{E}[V_k\mid U_k]$ is the client-wise Bayes predictor.  Let $\mathcal{H}$ be the hypothesis class in which the learned model $m$ lies. \\

\noindent\textbf{1. Decomposition via the best in-class predictor.}
Define the in-class oracle
\begin{equation}
    \tilde{m}_k \in \arg\min_{h\in\mathcal{H}} \mathbb{E}\!\left[\|\,m_k^\star(U_k)-h(U_k)\|_2^2\right].
\end{equation}

This is the (unique up to null sets) metric projection of the Bayes rule
onto the closed convex set $\{h(U_k):h\in\mathcal{H}\}\subset L^2$. By the Hilbert-space projection theorem, the residual $m_k^\star(U_k)-\tilde{m}_k(U_k)$ is orthogonal to the error $\tilde{m}_k(U_k) - m(U_k)$.  
Thus,
\begin{equation}
    \mathbb{E}\!\left[
        \langle\,
            m_k^\star(U_k)-\tilde{m}_k(U_k),\;
            \tilde{m}_k(U_k)-m(U_k)
        \rangle
    \right]
    =0,
\end{equation}
\emph{without requiring any unbiasedness assumptions}. Expanding the square and using this orthogonality yields the exact decomposition
\begin{equation}
    \begin{split}
        L_{\mathrm{approx}}^{(k)}(H,S;m)
        & = \underbrace{
            \mathbb{E}\!\left[\|\,m_k^\star(U_k)-\tilde{m}_k(U_k)\|_2^2\right]
            }_{\text{approximation (curvature) error}} \\
        & \quad +
        \underbrace{
            \mathbb{E}\!\left[\|\,\tilde{m}_k(U_k)-m(U_k)\|_2^2\right]
        }_{\text{estimation (variance) error}}.
    \end{split}
\end{equation} \\

\noindent\textbf{2. Bounding the curvature term.}
Assume the Bayes rule is twice differentiable on the intrinsic manifold
$\mathcal{M}(H)$, with uniformly bounded Hessian:
\begin{equation}
    \|\nabla^2 m_k^\star(u)\|_{\mathrm{op}} \le K_{2,k}, \quad
    \forall\,u\in\mathcal{M}(H).
\end{equation}

Let $K_2 := \max_k K_{2,k}$. Partition the intrinsic space into $N_k$ cells of diameter
$r_k\asymp N_k^{-1/d_{I,k}(H)}$. On each cell, the second-order Taylor remainder gives
\begin{equation}
    \|m_k^\star(u)-m_k^\star(u_0)\| \lesssim K_2 \, r_k^2,
\end{equation}
hence the best piecewise-constant or piecewise-affine approximation over cell diameter $r_k$ incurs squared error $\lesssim K_2^2 r_k^4$. Integrating over the manifold yields the curvature bound
\begin{equation}
    L_{\mathrm{curv}}^{(k)}(H) \le 
    C_{\mathrm{curv}}\ 
    K_2^2\ 
    d_{I,k}(H)^2\ 
    N_k^{-4/d_{I,k}(H)},
\end{equation}
for some constant $C_{\mathrm{curv}}>0$. \\

\noindent\textbf{3. Bounding the sampling variance term.}
Client $k$ has $D_k$ samples. Because overlapping windows reduce the number of statistically independent windows by a factor of $H$, the effective sample size is $\asymp D_k/H$. If the $N_k$ regions receive (on average) $\bar{n}_k \approx D_k/(H N_k)$ samples, standard nonparametric concentration arguments give
\begin{equation}
    L_{\mathrm{var}}^{(k)}(H,S) \le 
    C_{\mathrm{sample}}\ 
    \frac{d_{I,k}(H)\,H\,N_k}{D_k},
\end{equation}
for a constant $C_{\mathrm{sample}}>0$. (This term depends on $S$ only through the output dimension.) \\

\noindent\textbf{4. Optimizing over the resolution $N_k$.}
The total approximation loss satisfies
\begin{equation}
    \begin{split}
        L_{\mathrm{approx}}^{(k)}(H,S;m) 
        & \lesssim K_2^2 d_{I,k}(H)^2 N_k^{-4/d_{I,k}(H)} \\
        & \quad + \frac{d_{I,k}(H)\ H\ N_k}{D_k}.
    \end{split}
\end{equation}
Balancing the two terms in \(N_k\) yields
\begin{equation}
    \begin{split}
        L_{\mathrm{approx}}^{(k)}(H,S;m)
        & \lesssim \Big(K_2^2 d_{I,k}(H)^2\Big)^{\frac{d_{I,k}(H)}{4+d_{I,k}(H)}} \\
        & \quad + \Big(\frac{d_{I,k}(H)\,H}{D_k}\Big)^{\frac{4}{4+d_{I,k}(H)}}.
    \end{split}
\end{equation}

This gives the stated scaling of the approximation loss in terms of the
intrinsic dimension \(d_{I,k}(H)\), the horizon \(H\), and the available
client sample size \(D_k\).
\end{proof}

\end{document}